\documentclass[a4paper]{report}
\usepackage[utf8]{inputenc}
\usepackage[T1]{fontenc}
\usepackage{RJournal}
\usepackage{amsmath,amssymb,array}
\usepackage{algorithm}
\usepackage[noend]{algpseudocode}
\usepackage{booktabs}
\usepackage{xcolor}
\usepackage{wrapfig}
\usepackage{float}
\usepackage{amsthm}
\theoremstyle{definition}

\usepackage{amssymb}

\begin{document}

\sectionhead{MI2DataLab Preprint 
}
\volume{XX}
\volnumber{YY}
\year{20ZZ}
\month{AAAA}
\begin{article}

\title{Triplot: model agnostic measures and visualisations for variable importance in predictive models that take into account the hierarchical correlation structure} 

\author{by Katarzyna Pękala, Katarzyna Woźnica and Przemysław Biecek}

\maketitle

\abstract{
One of the key elements of explanatory analysis of a predictive model is to assess the \textbf{importance of individual variables}. Rapid development of the area of predictive model exploration (also called explainable artificial intelligence or interpretable machine learning) has led to the popularization of methods for \textbf{local (instance level) and global (dataset level) methods}, such as Permutational Variable Importance, Shapley Values (SHAP), Local Interpretable Model Explanations (LIME), Break Down and so on. However, \textbf{these methods do not use information about the correlation between features} which significantly \textbf{reduce} the explainability of the model behaviour.
In this work, \textbf{we propose new methods to support model analysis by exploiting the information about the correlation between variables}. 
The dataset level aspect importance measure is inspired by the block permutations procedure, while the instance level aspect importance measure is inspired by the LIME method. 
We show how to analyze groups of variables (aspects) both when they are proposed by the user and when they should be determined automatically based on \textbf{the hierarchical structure of correlations between variables.} 
Additionally, we present the \textbf{new type of model visualisation, triplot}, which  exploits a hierarchical structure of variable grouping to produce a high information density model visualisation. This visualisation provides a consistent illustration for either local or global model and data exploration.
We also show an example of real-world data with 5k instances and $37$ features in which a~significant correlation between variables affects the interpretation of the effect of variable importance.
The proposed method is, to our knowledge, the first to allow direct use of the correlation between variables in exploratory model analysis. 
Triplot package for R is developed under an open source GPL-3 licence and is available on the GitHub repository at \url{https://github.com/ModelOriented/triplot}.

}



\section{Introduction} \label{chap00_intro}




In the rapidly developing field of machine learning, we may notice two tendencies. On the one hand, we are observe growing complexity of predictive models, as well as increasing preference for the use of black-box solutions. Thanks to those, better performance may be assured on validation data, but at the cost of model interpretability. On the other hand, we need to acknowledge a growing need to understand where model results come from. Those tendencies contribute to the dynamic development of Explainable Artificial Intelligence (XAI) methods.

Model explanations provide multiple types of methods for assessing the importance of particular variables, extracting the character of the relationship between variable values and model predictions (e.g., Accumulated Local Effects and Partial Dependence),  identifying interactions (e.g., Friedman H-statistics), detection of model drift or residual diagnostic. 

Typically model analysis begins with the assessment of the importance of variables. Many XAI techniques have been developed to calculate the impact of a variable on model performance. Those methods can be model-specific - suitable only for one type of model (e.g., random forest of tree boosting) or model agnostic - work for many different types of models. Another frequently used categorization is between local techniques - focused on a single observation (e.g., Shapley values) and global techniques describing properties of the model on the whole population (e.g., aggregation of Shapley values).

A common property of the methods mentioned above is working for individual variables provided in data. But choosing to explain machine learning models by using variable importance for groups, instead of variable importance for single variables, may be useful in a few ways. 
Firstly, it may help to avoid obtaining misleading results because of correlated predictors. Those correlations are not necessarily immediately visible, but they may cause misleading explanations.

Secondly, such approach may help to reduce the size of the explanation by grouping similar variables together. It may prove to be helpful since an explanation of the model built on a highly dimensional dataset that contains several similar measures is not necessarily easy to interpret. For instance, let us consider here a FICO dataset, from The Explainable Machine Learning Challenge, which contains several possibilities of defaults in different time spans. Alternatively, we can think about a hypothetical model for the IOT dataset that contains multiple measurements for each parameter: temperature, humidity, or air quality. The contribution of those similar variables, such as defaults or temperature measurements, may be easier to interpret when shown as one number per variable type.

In the following article, we present a global and local, model agnostic approach to exploring explanatory variables correlations, creating groups of variables and calculating their importance to the predictions of machine learning models. Those groups may contain correlated, or in other ways similar, variables.

\subsection {Variable importance}

There is a number of R packages that provide tools for calculating model-agnostic variable importance. Table \ref{table:pckg_downloads} presents some of them, with additional information whether the delivered explanation is local or global. We notice that most of the packages have been developed very recently.

\begin{table}[H]
    \begin{center}
    \begin{tabular}{ l c c c c c } 
    \hline
    \textbf{package}  &\textbf{downloads} & \textbf{date published} & \textbf{age in months} & \textbf{local} & \textbf{global} \\ 
    \hline
     triplot & 3548 &  2020-06-09 & 9 & \checkmark  & 
     \checkmark \\ 
     ExplainPrediction & 27680 &  2015-09-07 & 66  & \checkmark  & 
     \checkmark \\ 
     DALEX & 100780 & 2018-02-28 &  36 & \checkmark  &  \checkmark \\ 
     iml  & 127400 & 2018-03-13 &  36  &  \checkmark  &  \checkmark \\  
     lime   & 110722 & 2017-09-15 &  42 &  \checkmark &   \\ 
     localModel     & 10353 & 2019-04-14 &  23 &  \checkmark &   \\ 
     vimp & 14919  & 2018-06-24 &  33 &   &  \checkmark \\ 
     vip & 120133 & 2018-06-15 &  33 &   &  \checkmark \\  
    \hline
    \end{tabular}
    \caption{The table presents a list of CRAN packages that provide model-agnostic explainers for variable importance, with additional information on how many times they were downloaded, the date of first publication, how many months passed since they were published, and whether they calculate local and/or global statistics. Data for this table were gathered on March 27, 2021.}
    \label{table:pckg_downloads}
    \end{center}
\end{table}

The \CRANpkg{ExplainPrediction} \citep{Robnik-Sikonja_Kononenko_2008} package provides local and global explanations for different classification and regression models. The \CRANpkg{DALEX} package \citep{JMLRv19} supports local and global model agnostic measures of variable importance. For global explanations, DALEX uses a function \code{model\_parts} that is a model agnostic method for calculating the permutation-based feature importance. Additionally, \code{model\_parts} faciliates calculation of the  contribution of variable groups. The \CRANpkg{iml} package \citep{molnar2020} provides the whole set of tools for measuring global and local level feature importance. Packages \CRANpkg{lime} \citep{lime2019} and \CRANpkg{localModel} \citep{localmodel2019} both implement the LIME method of explaining predictions. The package \CRANpkg{vimp} \citep{vimp2020} provides a model-level variable importance measure with point and confidence interval estimates. It also calculates grouped variable importance. The package \CRANpkg{vip} \citep{vip2020} contains several functions (i.e., Shapley values) for calculating global variable importance.


Finally, the R package \CRANpkg{RFgroove}, developed in 2015, delivered the \code{varImpGroup} method and it is also worth noting. It allowed to calculate the permutation variable importance for Random Forest; specifically it facilitated computations for the importance of groups of variables. The package is now archived.

\subsection {Importance of group of variables}
Calculating the importance of single variables prediction is very helpful in discovering how the black-box works. Unfortunately, many methods do not take into account the structure of the correlation between the analyzed predictors. This can lead to incorrect conclusions. To prevent this, we can calculate the importance of the groups of correlated predictors, thus avoiding misleading results.

Additionally, measuring the importance of the groups of variables may support the effort of model building in a couple of ways: as described in \cite{gregorutti2015}, it may be useful in increasing the model interpretation and help increase prediction accuracy. In addition, it may also support the feature selection effort. 

Model agnostic importance for groups of variables is not widely implemented. Here, we will use \pkg{DALEX} and \pkg{triplot} functions that allow calculation of feature importance for the groups of variables that works regardless of the used model.

\subsection {Feature importance and correlation}

We can approach building groups of variables in multiple ways. We can use the Variance Inflation Factor (VIF), which is a tool for discovering multicollinearity in linear models \citep{james2013}. Alternatively, we can use correlation and analyze pair-wise correlations in form of the correlation matrix. R packages offer many tools to visualize such a matrix, for example function \code{corrplot} from package \CRANpkg{corrplot} \citep{corrplot2017}. However, when we need to analyze groups of correlated variables that are bigger than 2 elements, the matrix is not a sufficient tool for that. In this case, we can use packages that visualize the correlation between variables in the form of a network or a graph. 

The \CRANpkg{Corrr} \citep{corrr2020} package provides a network plot as a method of visualizing correlation. By keeping correlated variables in clusters and using colors to denote correlation strength, it facilitates exploration of the correlation structure. 
The recently developed package \CRANpkg{corrgrapher} \citep{corrgrapher2020} shows the correlation in the form of a graph that can be interacted with. Edge lengths between variables are determined by the strength of their correlation. By inspecting such network as provided by corrr, or graph drawn by corrgrapher, we can visually assess possible groupings of correlated variables.

Another solution is to create a hierarchical clustering tree with correlation as a distance measure. Such a tree can be created by using the package \CRANpkg{cluster} \citep{cluster2019} and method agnes (agglomerative hierarchical clustering). Plotting the tree allows us to group variables according to their pair-wise correlations. 

\begin{figure}
  \begin{center}
    \includegraphics[width=0.42\textwidth]{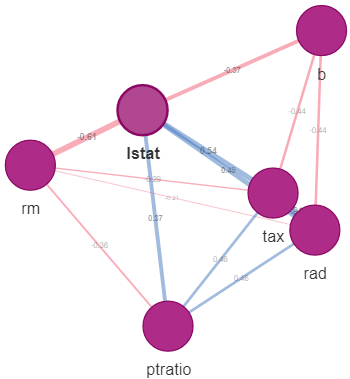}
  \end{center}
  \caption{Exploring dataset structure with corrgrapher - showing pair-wise correlation for 6 variables, from the BostonHousing2 dataset from the \CRANpkg{mlbench} package \citep{mlbench2012}.}
  \label{figure:Boston_corrgrapher}
\end{figure}


\section{Global and local variable importance}\label{chap_VI}
In the following chapter we describe methods for assessing the variable importance at the model and instance levels used in \CRANpkg{triplot}.

\subsection{Global variable importance}

Permutation variable importance, used in the \pkg{triplot} package, facilitates calculation of the importance of explanatory variables in a given machine learning model, across the whole dataset \citep{fisher2019}. 

It works by permuting a single variable and calculating how much loss in performance of the model this permutation causes. A significant loss means that the permuted variable is important for the model. The method also facilitates measurement of the importance of the group of variables, by measuring the loss after permuting the whole group of them.


\subsection{Local variable importance - introducing the predict aspects method}

There exist a number of methods for local explanation of black-box models, like Break Down, Shap or LIME. However, there is one disadvantage common to these solutions - they do not take into account the correlations of explanatory variables.

The pedict aspects method aims to increase the interpretability of the model by providing an instance-level explainer for the whole groups of explanatory variables. It enables grouping predictors and calculates the contribution of those groups to the prediction. As a result, we increase the readability of the explanation by reducing the number of components contributing to the prediction. Additionally, we can also acknowledge the occurrence of correlated variables by placing them in one group.

\textbf{Method}

\begin{figure}[H]
      \includegraphics[width=\textwidth]{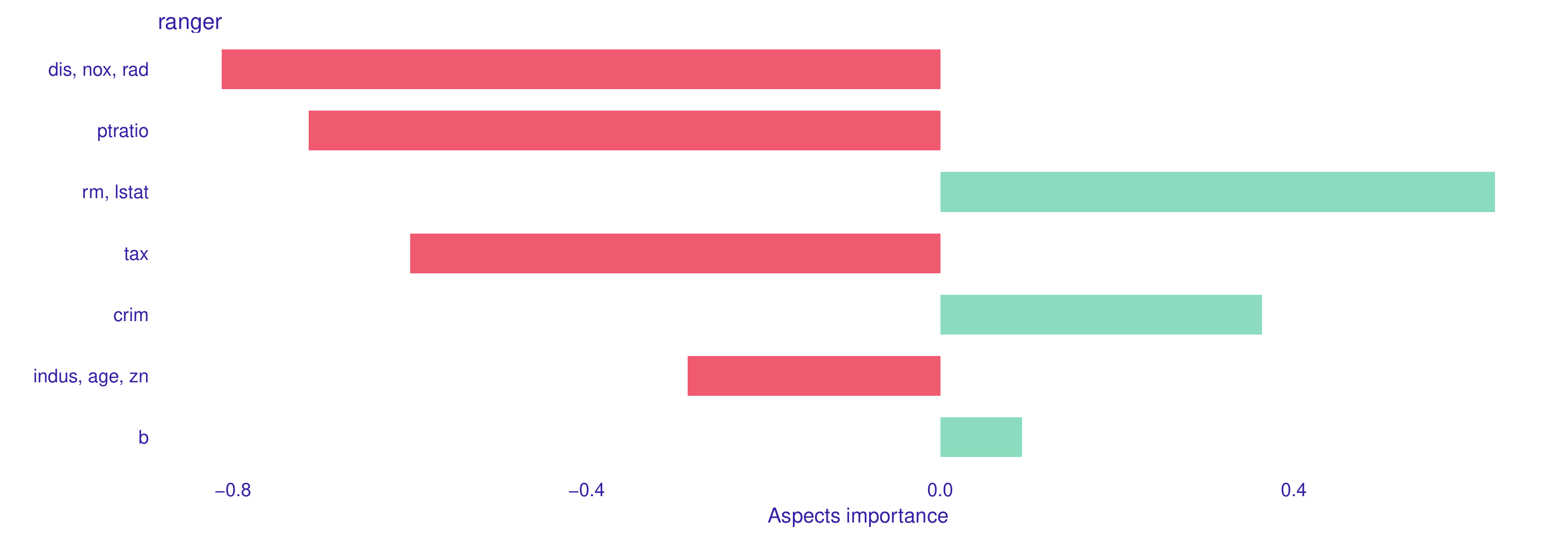}
      \caption{ \code{predict\_aspects} results show the group of variables contribution to the chosen, single prediction, for the BostonHousing2 dataset from the \pkg{mlbench} package.}
      \label{figure:predict_aspects}
\end{figure}

Our goal is to understand how groups of variables contribute to the calculated prediction of a chosen observation in the machine learning model. Those groups, called aspects, can be built automatically (by joining together correlated variables) or manually, based on prior knowledge.  Afterwards, we calculate the contribution to the prediction of every group of predictors (aspects). Hence the method predicts aspects. Results can be plotted, as presented in Figure \ref{figure:predict_aspects}.

The function \code{$predict\_aspect$} was inspired by the LIME method \citep{ribeiro2016why}. This a well-known XAI approach, which explains black boxes by building interpretable models locally, around the prediction. In the case of sparse datasets like images, a new dataset is built by splitting the image into super-pixels and perturbing them by randomly graying them out. The interpretable model is built on this new dataset. 

In the case of the \code{$predict\_aspect$} method we are working on tabular data. A new dataset is built by subsetting observations from the original dataset and then modifying them, so every sampled observation will have at least one aspect replaced by the data from the observation of interest. Then a linear regression model is built. Its coefficients predict how those replacements change the prediction of the modified data. A detailed procedure of computing \code{$predict\_aspect$} is described in Algorithm \ref{alg:predict_aspect}.

\begin{algorithm}
\caption{Predict aspects}\label{pa_algorithm}\label{alg:predict_aspect}
\begin{algorithmic}[1]
\Procedure{}{}
\State $\mathcal{X}$ - dataset of size $n \times p$
\State $f$ - model built on $\mathcal{X}$
\State $x_*$ - observation to be explained   
\State $\mathcal{P} \gets$ group explanatory variables into aspects, $\mathcal{P} = {q_1, ..., q_m}$, partition of set of indexes $J = {1, ..., p}$  
\State $A \gets$ sample $N$ observation from $\mathcal{X}$, with replacement
\State $X' \gets$ zero matrix of size $N \times m$
\For {every row of $X'$}
    \State sample, with replacement, two columns indexes $k$,$l$
    \State replace $0$ with $1$ at the given row's and column's $k$ and $l$ intersection of $X'$
\EndFor
\State creating $A'$ by taking $A$ and replacing chosen observations by data from $x_*$, replacement is directed by $X'$
\State calculate $Y_m = f(A') - f(A)$  
\State fit linear model $g$, $g(X') = Y_m$
\EndProcedure
\end{algorithmic}
\end{algorithm}

Since we fit the linear model, the coefficients vector can be expressed precisely. We assume that $Y_m = \gamma X' + \varepsilon$, where $\gamma$ is a vector of coefficients. As solution of least square regression
$$\gamma = (X'^{T} X')^{-1} X'^T Y_m.$$
Let us denote as $W = X'^{T} X'$ and $Z = X'^T Y_m$. Additionally, let us define a set of auxiliary functions $\sigma_j (i) = x[i, j]$, for $j = 1, \dots m$ and $i = 1, \dots n$.  So the function $\sigma_j$ is an indicator that $j$ aspect is replaced for $i$-th observation in matrix $X$.
We can express matrices $W$ and $Z$ in terms of $sigma$ functions. The values of matrix $W$ are equal to
$$w[i, j] = \begin{cases} 
\sum_k \sigma_i (k) * \sigma_j (k), & \text{for } i \neq j \\
\sum_k \sigma_i (k),  & \text{for } i = j
\end{cases} $$ and count how many single aspects or combinations of two aspects are modified. 
Vector $Z$ is the difference between modified and unmodified predictions for a corresponding $i$-th aspect
\begin{equation}
\label{eq:z_vector}
    z[i] = \sum_k \sigma_i (k) * (f(a_k') - f(a_k)).
\end{equation} 


Looking at the equation \ref{eq:z_vector} we can observe that some of the values may decrease if differences in predictions cancel each other out for the given aspect. Then, the effect of that individual aspect may be disregarded in estimated coefficients.

If  $x_*$ is \textit{an average observation}, most $Z$ vector's values will be close to $0$.

It i worth noting that before we can use the method, we need to group the explanatory variables into aspects. We can use two different approaches: we can build the aspect list arbitrarily or we can use the \code{group\_variables} function that will do the grouping for us by using variables correlations. In the second approach, we are going to get aspects where every absolute value of the pair-wise correlation of explanatory variables is no smaller than a given level. It should be noted that \code{group\_variables} works only for numerical variables.

\subsection{Controlling the number of explained variables with lasso}

To further increase our understanding of the model, we can use a similar approach as in LIME - limit the number of contributing aspects by using the predict aspect method with lasso. It is especially useful when the model is built on highly dimensional data. Using lasso allows to control how many aspects have a non-zero contribution value. This can prove useful when, after the first round of analysis, we find out that there are many groups of variables that have a very small contribution to the prediction. By using lasso regression we can remove them from the final explanation. 

\begin{figure}[H]
      \includegraphics[width=\textwidth]{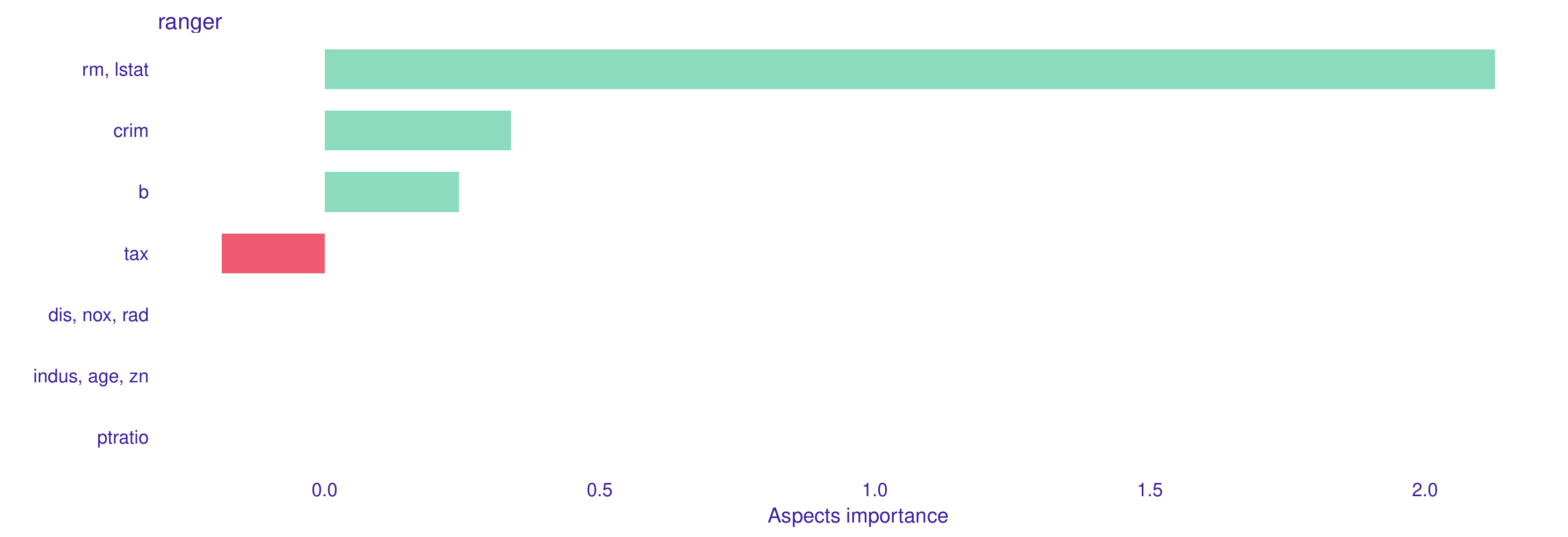}
      \caption{\code{predict\_aspects} results show the group of variables contribution to the single prediction, after using lasso regularization, for the BostonHousing2 dataset from the \pkg{mlbench} package.}
      \label{figure:predict_aspects_lasso}
\end{figure}

If we decide that the explanation should have a specific, limited number of nonzero aspects, the \code{predict\_aspects} method can use the lasso regularization in its final stage, instead of building a standard linear model. More specifically, linear regression with lasso is executed multiple times on the binary matrix $X'$. Afterwards, the smallest possible regularization parameter lambda is selected in such a way that it guarantees keeping the chosen limit. In the end, it extracts coefficients (contribution values) after regularization with a given lambda.
The model explained in Figure \ref{figure:predict_aspects} is shown after using lasso in Figure \ref{figure:predict_aspects_lasso}.

\section{Understanding variables importance with triplots}\label{chap_triplots}
In this chapter, we introduce the triplot tool that enables exploration of the importance of correlated variables. 

\subsection{Motivation}

The aim of triplots is to evaluate the importance of the variables in a multi-aspect approach, taking into account the correlation structure. The variable importance analysis can be performed at the whole dataset level (global level) as well as at the single prediction level (local level).

Triplots may be especially useful when we are not fully familiar with the internal data structure and we are not able to decide in advance on the good separation of variables into groups.

\subsection{Understanding triplots}

Triplots show in one place:
\begin{itemize}
\item the importance of individual variables (left panel),
\item global correlation structure visualized by hierarchical clustering (right panel),
\item the importance of groups of variables determined by hierarchical clustering (middle panel).
\end{itemize}

The order of aspects determined by hierarchical clustering allows us to check the values of variables' importance for the different levels of variables grouping. 

Demonstrating in one plot the change of variables' importance that stems from modification of aspects sizes, facilitates a deeper understanding of how correlation influences the variable's single contribution to the prediction. Based on these findings, we can determine the best approach to variable grouping and decide how to further develop the model explanation.

\begin{figure}[H]
      \includegraphics[width=\textwidth]{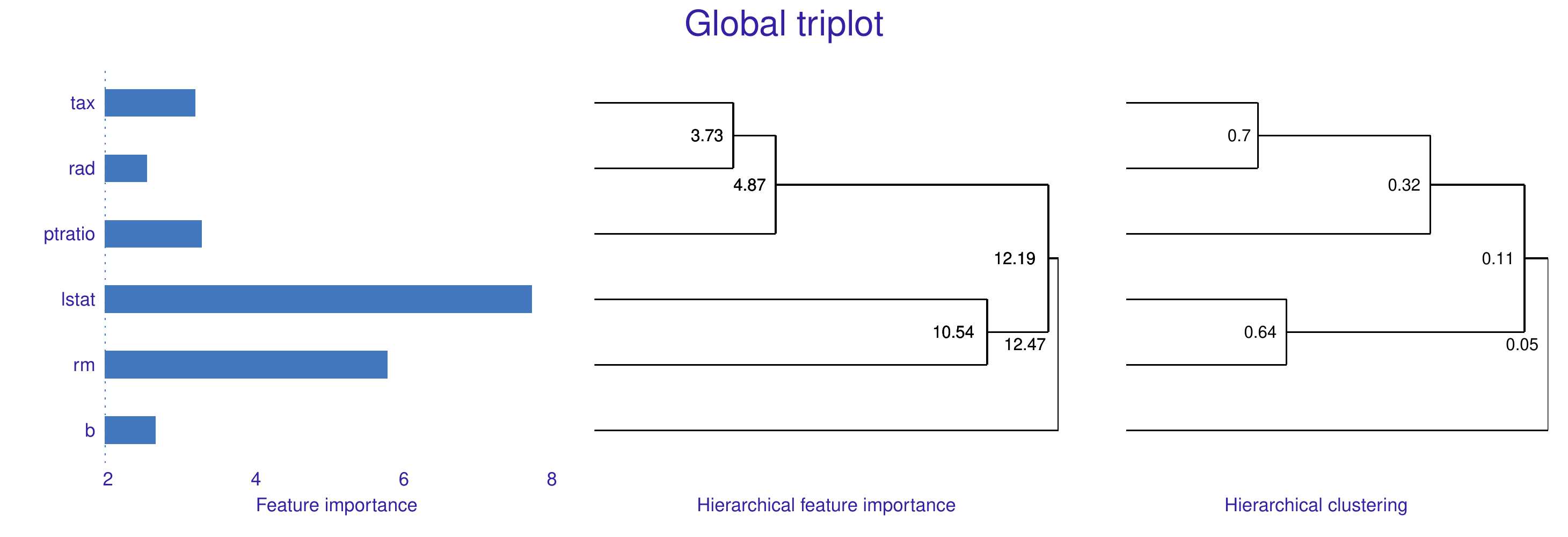}
      \caption{Global triplot for 6 variables from the BostonHousing2 dataset included in the \pkg{mlbench} package}
      \label{figure:global_triplot_6}
\end{figure} 

\subsection{Triplots methodology}

Calculating triplot starts with running a variable importance algorithm on the dataset and calculating the contribution of each single explanatory variable. When the analysis is to be done at a global level (global triplot), triplot uses \code{model\_parts} implementation of a variable importance algorithm, included in  the \pkg{DALEX} package. Otherwise, for the local level calculations (local triplot), triplot is applying its \pkg{triplot's} package \code{predict\_aspects} method to evaluate the contribution. 

Next, triplot calculates the hierarchical clustering tree. For dissimilarity measure it uses Spearman's correlation by default and the linkage method is of a \textit{complete} type (it allows usage of other linkage methods). As a result, we get the order in which we join features. Figure \ref{figure:global_triplot_6} presents an example of a global triplot. 

It should be noted that \code{calculate\_triplot} works for datasets with only numerical variables.

\begin{algorithm}
\caption{Calculating triplot}\label{tri_algorithm}
\begin{algorithmic}[1]
\Procedure{}{}
\State $p\gets$ number of predictors
\State Calculate hierarchical clustering tree $T$
\State $idn\gets$ list of $T$ nodes that contain aspects components
\For {$i$ in ${1,...,p} $}
    \If {triplot global}
        \State Calculate $vip_i$
    \Else
        \State Calculate $prediction\_aspects_i$
    \EndIf
\EndFor
\For {$i$ in nodes}
        \If {triplot global}
        \State Calculate $vip_i$
        \State Where $j$ set of indices from $idn$
    \Else
        \State Calculate $prediction\_aspects_i$
        \State Where $j$ set of indices from $idn$
    \EndIf
\EndFor
\If {triplot global}
    \State Calculate baseline 
\EndIf
\EndProcedure
\end{algorithmic}
\end{algorithm}

\newpage


\newpage

\section{Example on the FIFA dataset} \label{chap_demo}
In the following example, we look at the Fifa dataset that contains the specification of football players. The analysis starts with the exploration of the internal data structure with \pkg{corrplot} and \pkg{corrgrapher}. Next, we build a random forest model to predict the players' value in Euro. In order to explain the model, we use the \code{model\_triplot} function from the \pkg{triplot} package to examine the global variables' importance. Afterwards, we choose one player and we focus on explaining variables' contribution to the prediction of his value. 

The Fifa dataset is available in the \pkg{DALEX} package. After importing the dataset, we introduce several modifications before further work.

\begin{example}
library("DALEX")
data(fifa)
fifa$value_eur <- fifa$value_eur/10^6
fifa[, c("nationality", "overall", "potential", 
         "wage_eur")] <- NULL

\end{example}

\begin{figure}[ht]
      \includegraphics[width=\textwidth]{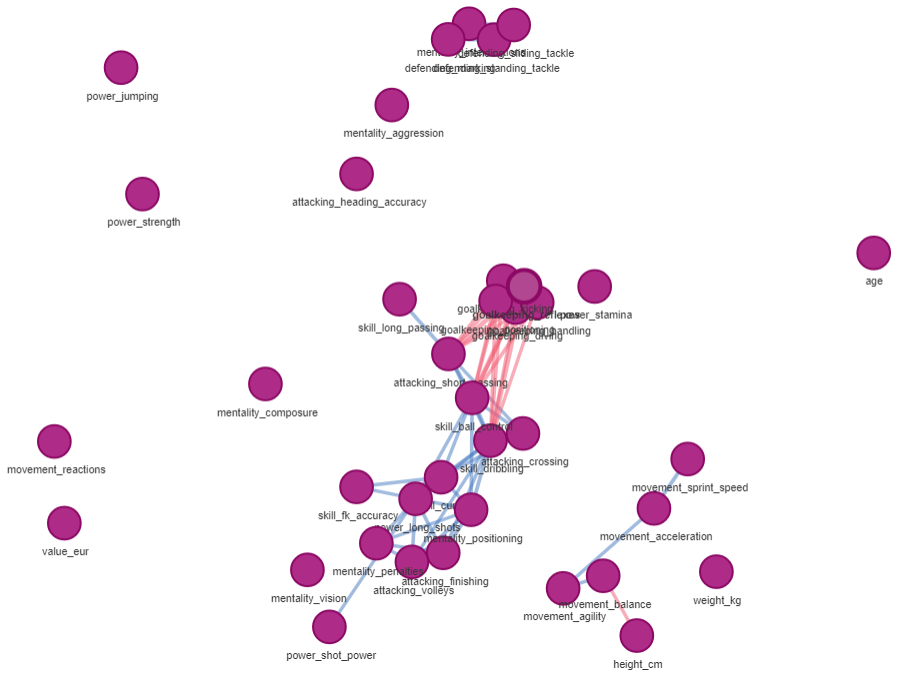}
      \caption{Corrgrapher for the fifa dataset with the cut off point at 0.5.}
      \label{figure:fifa_corrgrapher}
\end{figure}

To investigate the dataset's correlations we use a tool from the package \pkg{corrplot} that allows us to create a colored correlation matrix as presented in Figure \ref{figure:fifa_corrgrapher}.

\begin{example}
library("corrplot")
fifa_corr_mat <- cor(fifa)
corrplot(fifa_corr_mat, method = "color",
         type = "upper", order = "hclust", 
         addCoef.col = "black", number.cex = .45, tl.cex = 0.8,
         diag = FALSE)
\end{example}

\begin{figure}[ht]
      \includegraphics[width=\textwidth]{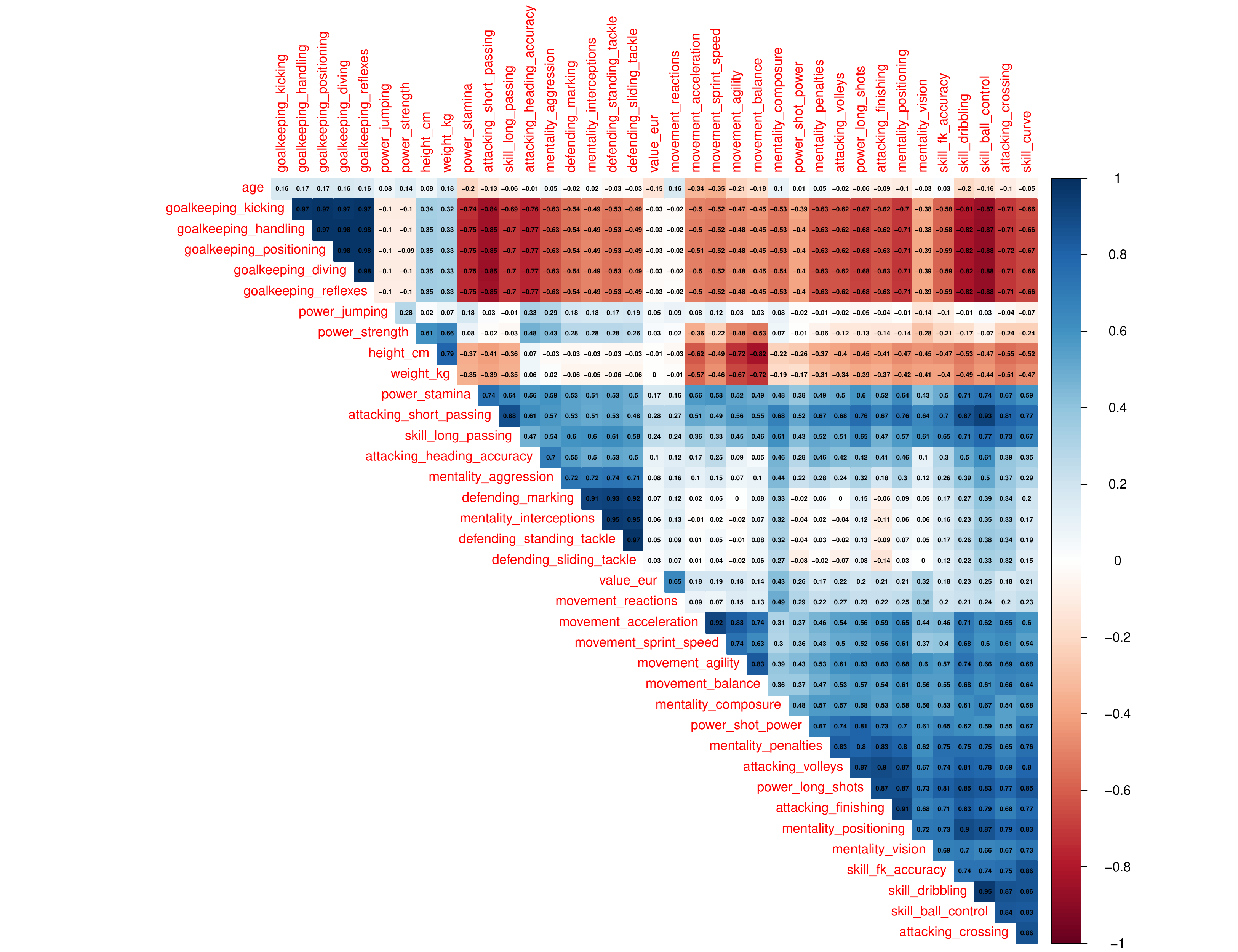}
      \caption{Corrplot for the fifa dataset.}
      \label{figure:fifa_corrplot}
\end{figure} 

We can observe some groups of highly correlated variables like, e.g.: 

\begin{itemize}
\item goalkeeping skills and short passing,
\item defending skills and interceptions,
\item acceleration and sprint speed,
\item agility, balance and height,
\item positioning, penalties, volleys, finishing and long shots,
\item curve, dribbling, ball control and crossing.
\end{itemize}

We continue the correlation analysis by creating a corrgrapher object for the dataset.

\begin{example}
library("corrgrapher")
corrgrapher(fifa, cutoff = 0.8)
\end{example}

As we see in Figure \ref{figure:fifa_corrgrapher}, corrgrapher shows, in the form of a graph, pair-wise variables correlation. While exploring it, we can clearly distinguish several groups of correlated variables, in a slightly different manner than in case of a corrplot. 

After recognizing the correlations of the variables, we build a model with the \CRANpkg{ranger} package. We use \code{explain} from the \pkg{DALEX} package, on top of which we can easily create any explanations that are available in the DALEX toolbox. 

\begin{example}
library("ranger")
set.seed(2020)
fifa_model <- ranger(value_eur~., data = fifa)
fifa_explainer <- DALEX::explain(fifa_model,
                                 data = fifa[,-1],
                                 y = fifa\$value_eur,
                                 label = "Random Forest", 
                                 verbose = FALSE)
\end{example}

On the created explainer, we build and plot a global triplot.

\begin{example}
library("triplot")
fifa_triplot_global <- model_triplot(fifa_explainer, 
                                     B = 1, N = 5000,
                                     cor_method = "pearson")
plot(fifa_triplot_global, margin_mid = 0)
\end{example}   

\begin{figure}[ht]
      \includegraphics[width=\textwidth]{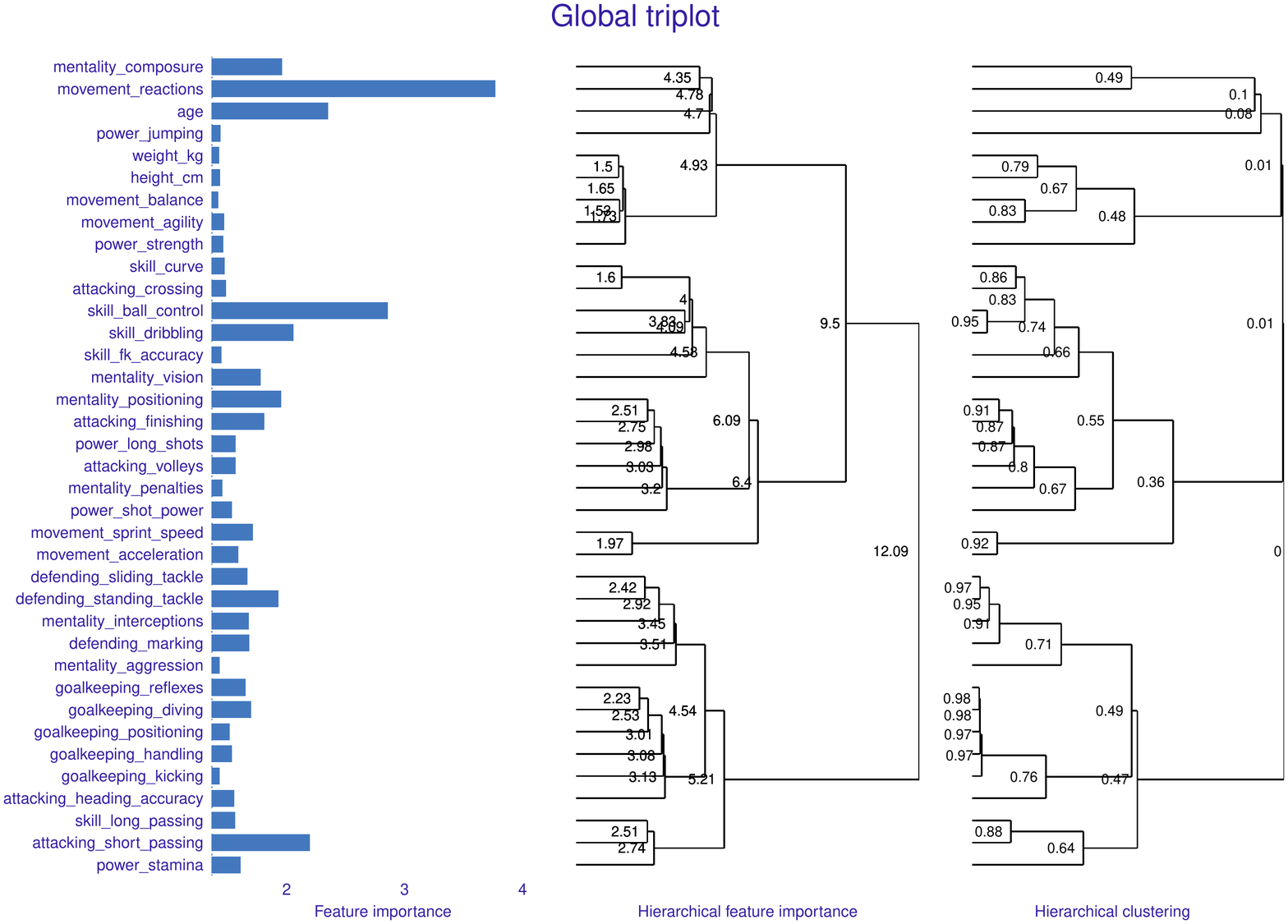}
      \caption{Global triplot for the fifa dataset }
      \label{figure:fifa_tri_global}
\end{figure}

As we can see in Figure \ref{figure:fifa_tri_global}, ball control and dribbling are strongly correlated (0.95) and a group formed out of them has the importance of 3.83.
Variables: (mentality) positioning, finishing, volleys, penalties, long shots and shot power are correlated (minimum pair-wise correlation is 0.67), but as a group, they have importance at only 3.2.
We can observe that reactions are the most important single variable. Together with composure, it forms the only two-element group of somewhat correlated variables with relatively high importance. Neither reactions nor composure are correlated with the other explanatory variables in this dataset. 

\begin{table}[H]
\small
    \begin{center}
    
    \begin{tabular}{  m{2cm} m{10cm} } 
    \hline
    \textbf{category}  &\textbf{skill} \\
    \hline
        age & age \\ \hline
        body & height\_cm, weight\_kg \\ \hline
        attacking & attacking\_crossing, attacking\_finishing, attacking\_heading\_accuracy, 
        attacking\_short\_passing, attacking\_volleys \\ \hline
        skill & skill\_dribbling, skill\_curve, skill\_fk\_accuracy, skill\_long\_passing, skill\_ball\_control \\ \hline
        movement & movement\_acceleration, movement\_sprint\_speed, movement\_agility, movement\_reactions, movement\_balance \\ \hline
        power & power\_shot\_power, power\_jumping, power\_stamina, power\_strength, power\_long\_shots \\ \hline
        mentality & mentality\_aggression, mentality\_interceptions, mentality\_positioning, mentality\_vision, mentality\_penalties, mentality\_composure \\ \hline
        defending & defending\_marking, defending\_standing\_tackle, defending\_sliding\_tackle \\ \hline
        goalkeeping & goalkeeping\_diving, goalkeeping\_handling, goalkeeping\_kicking, goalkeeping\_positioning, goalkeeping\_reflexes \\ \hline
    \end{tabular}
    
    \caption{Table presents the split of skills between categories.}
    \label{table:skills_split}
    
\end{center}
\end{table}

Now we are going to explain predictions for single players. Triplot presents explanations for models where aspects are built on variables correlations. For an explanation of single observation prediction, we are going to use grouping based on variables' category, instead of their correlation. Aspects' composition is shown in Table \ref{table:skills_split}.

\begin{example}

fifa_aspects <- list(
  "age" = "age",
  "body" = c("height_cm", "weight_kg"),
  "attacking" = c("attacking_crossing",
                  "attacking_finishing", "attacking_heading_accuracy",
                  "attacking_short_passing", "attacking_volleys"),
  "skill" = c("skill_dribbling",
              "skill_curve", "skill_fk_accuracy", "skill_long_passing",
              "skill_ball_control"),
  "movement" = c("movement_acceleration", "movement_sprint_speed",
                 "movement_agility", "movement_reactions", "movement_balance"),
  "power" = c("power_shot_power", "power_jumping", "power_stamina", 
              "power_strength", "power_long_shots"),
  "mentality" = c("mentality_aggression", "mentality_interceptions",
                  "mentality_positioning", "mentality_vision", 
                  "mentality_penalties", "mentality_composure"),
  "defending" = c("defending_marking", "defending_standing_tackle",
                  "defending_sliding_tackle"),
  "goalkeeping" = c("goalkeeping_diving",
                    "goalkeeping_handling", "goalkeeping_kicking",
                    "goalkeeping_positioning", "goalkeeping_reflexes"))

\end{example}

For the analysis we choose the best player. 

\begin{example}

top_player <- fifa[order(fifa$value_eur, decreasing = TRUE),][1,]
top_player$value_eur
[1] 105.5

fifa_explainer$y_hat[order(fifa$value_eur, decreasing = TRUE)[1]]
[1] 89.91145

\end{example}

We see that the prediction for the chosen observation, calculated by the model, is too low. For explaining which variables contribute to it, we use the \code{predict\_aspects} function and an aspect list that we have already created.

\begin{example}
fifa_aspects_importance <- predict_aspects(fifa_explainer, 
                                           new_observation = top_player, 
                                           variable_groups = fifa_aspects)
plot(fifa_aspects_importance)
\end{example}

\clearpage

\begin{figure}[h!t]
      \includegraphics[width=\textwidth]{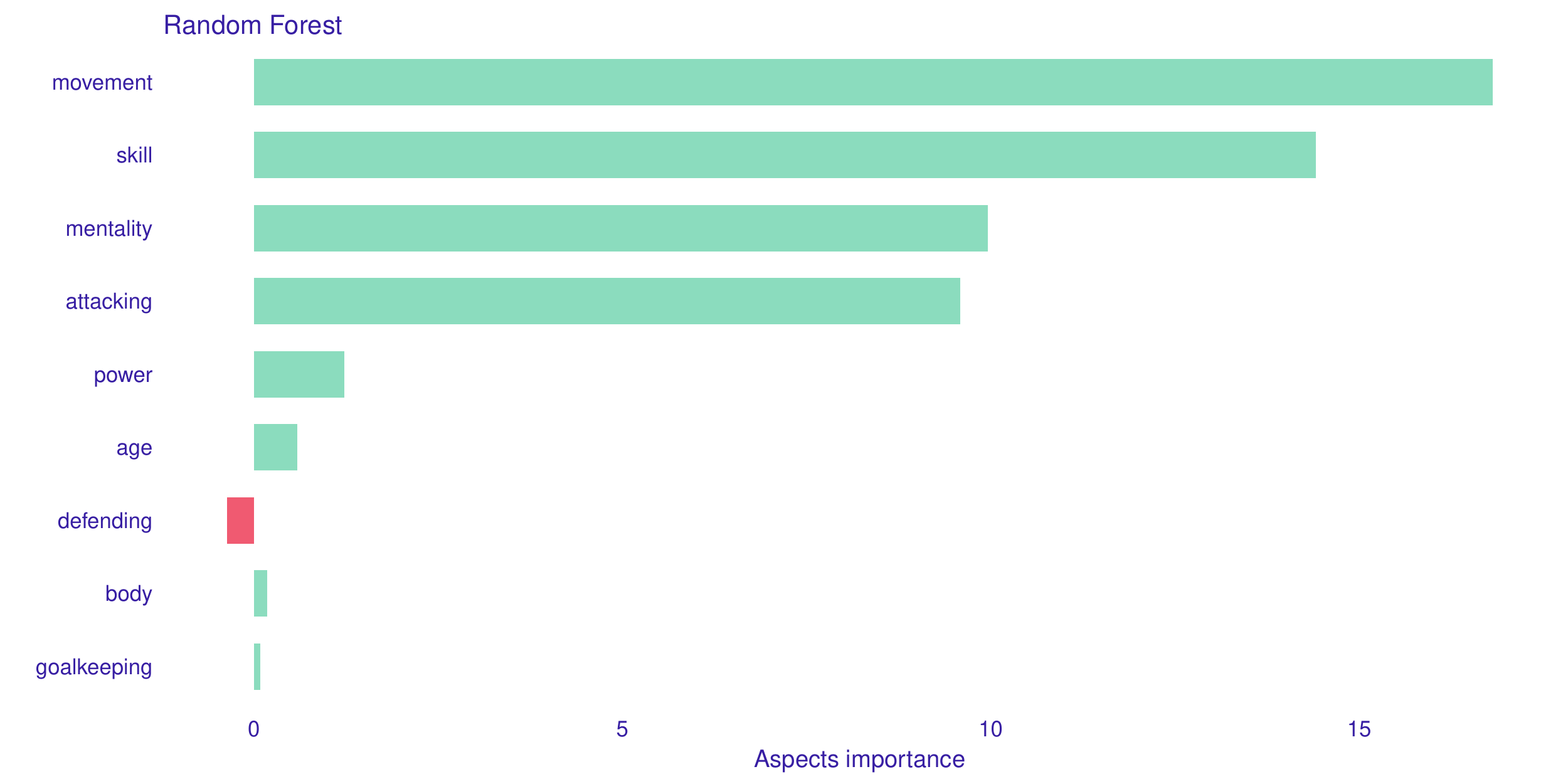}
      \caption{Contribution of variables' groups to chosen player's prediction, the fifa dataset }
      \label{figure:fifa_predict_aspetcs}
      \includegraphics[width=\textwidth]{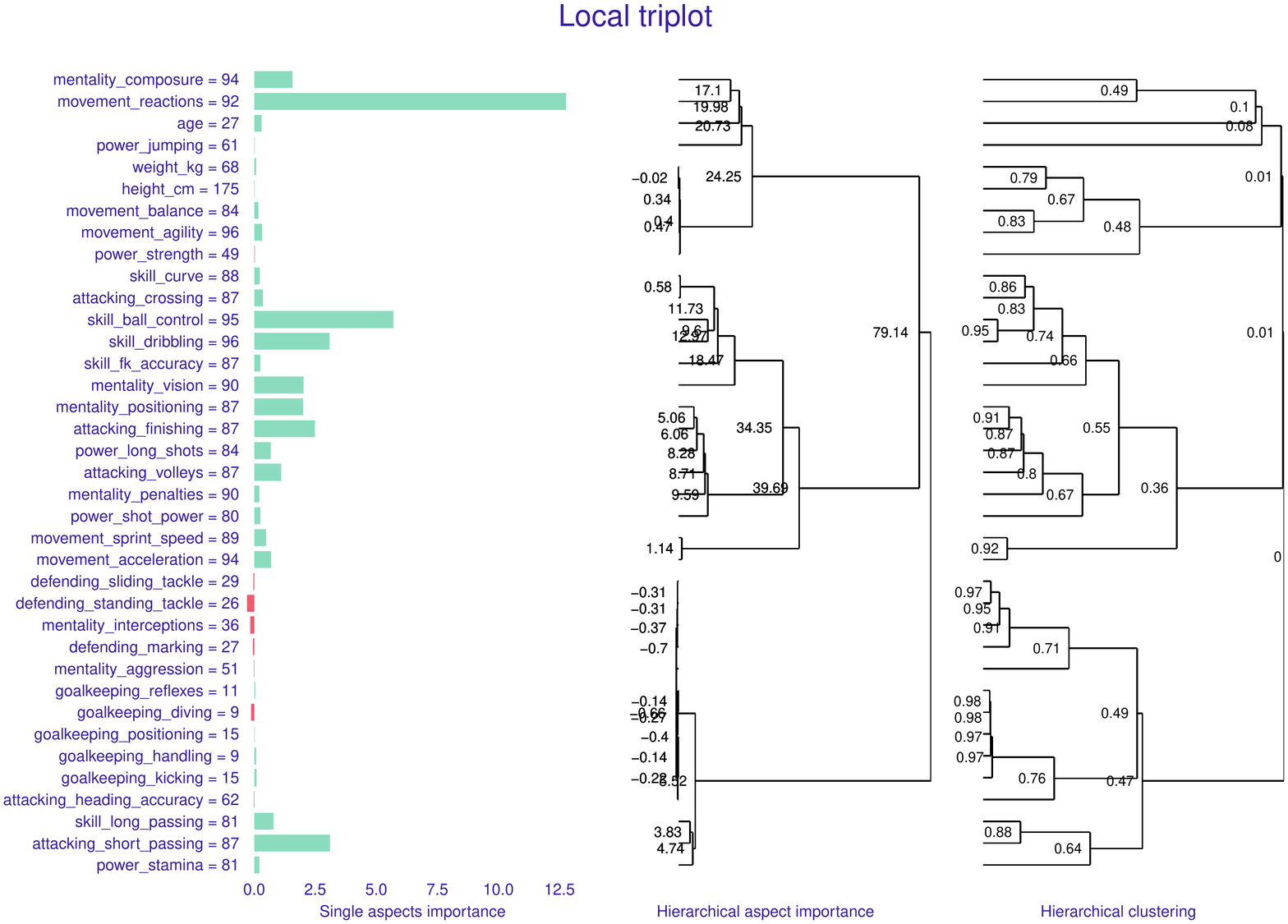}
      \caption{Local triplot for the fifa dataset }
      \label{figure:fifa_tri_local}
\end{figure} 
 
In Figure \ref{figure:fifa_predict_aspetcs}, we can observe that for the chosen player, variables grouped in aspects: movement, skill, mentality, attacking have a significant, positive contribution to the player's value prediction. The other variables have a much smaller impact. Defending has negative (albeit small) influence on the prediction.

The above explanation is created for only one grouping of variables, done manually. To explore explanations for automated, multiple groupings of variables, we use local triplot. 

\begin{example}
fifa_triplot_local <- predict_triplot(fifa_explainer, top_player,
                                      N = 5000, 
                                      cor_method = "pearson")
plot(fifa_triplot_local)
\end{example}

Figure \ref{figure:fifa_tri_local} shows that several groups of correlated variables have no significant contribution to the prediction (defending and goalkeeping skills, aggression, interceptions, as well as body parameters, balance, agility and strength). 

By investigating correlations in the triplot's right panel and strength of aspects' contributions in the middle and left panels, we can choose the appropriate way of grouping variables in the final explanation.


\section{Conclusion and future work} \label{chap_conclusions}
In this article, we described possible challenges posed by explaining the machine learning black boxes with single variables' importance. We presented a method of exploring grouped variables importance at the global and local level, as a way to overcome challenges that come from variable correlations. We introduced an experimental function \code{predict\_aspects} that allows analysis of local grouped variable importance.

The disadvantage of the \code{predict\_aspects} method is the lack of stability in case of low $N$  value. This problem could be approached in the future in a few ways. The method could be run a number of times and the final prediction could be calculated as an average of the following iterations. Alternatively, \code{predict\_aspects} could be based on BreakDown or Shapley methods, instead of on LIME. This could produce more stable results.

\section{Acknowledgements}

The Work on this paper is financially supported by the NCN Opus grant 2017/27/B/ST6/01307.

\bibliography{pekala_woznica_biecek}

\address{Katarzyna Pękala\\}
\email{katarzyna.pekala@gmail.com}

\address{Katarzyna Woźnica\\
Faculty of Mathematics and Information Science\\
  Warsaw University of Technology\\
  Poland\\}
\email{k.woznica@mini.pw.edu.pl}

\address{Przemysław Biecek\\
  Faculty of Mathematics, Informatics and Mechanic\\
  University of Warsaw\\
  Poland\\
  ORCID: \url{https://orcid.org/0000-0001-8423-1823} \\}
\email{przemyslaw.biecek@gmail.com}

\end{article}

\end{document}